\documentclass[runningheads]{llncs}

\usepackage[
  width=120mm,
  left=13.5mm,
  paperwidth=146mm,
  height=193mm,
  top=12mm,
  paperheight=217mm
]{geometry}

\usepackage{graphicx}
\usepackage{booktabs}
\usepackage{amsmath,amssymb}
\usepackage{array}
\usepackage{xcolor}
\usepackage{placeins}

\usepackage{makecell}

\usepackage{caption}
\usepackage[hidelinks]{hyperref}
\providecommand{\Cref}[1]{\autoref{#1}}
\providecommand{\cref}[1]{\autoref{#1}}

\newcommand{\equalcontrib}{\textsuperscript{\normalfont †}}
\newcommand{\corresponding}{\textsuperscript{*}}


\begin{document}

\title{Structure-Enhanced Features and Quality-Aware Dynamic Anchor Scoring for Robust Lane Detection
}
\titlerunning{Structure-Enhanced Features and Quality-Aware Dynamic Anchor Scoring
}

\author{Weize Cai\inst{1}\textsuperscript{,\equalcontrib} \and Yongqi Dong\inst{2}\textsuperscript{,\equalcontrib,\corresponding} \and Zhida Shao\inst{1} \and Yichen Liu\inst{1} \and Zixin Fu\inst{3}}  
\authorrunning{W. Cai et al.}

\institute{
RWTH Aachen University, Aachen, Germany
\and
Delft University of Technology, Delft, The Netherlands\\
\and
Chang'an University, Xi'an, China \\[3pt]
\textsuperscript{†} Equal contribution. \quad
\textsuperscript{*} Corresponding author: \email{yongqi.dong@rwth-aachen.de}. 
}

\maketitle

\begin{abstract}
Lane detection requires recovering thin, elongated, and frequently occluded lane structures under challenging driving conditions. While anchor-based detectors provide efficient candidate generation, their performance is limited by two coupled issues: backbone features often lose structural continuity along partially visible lanes, and classification confidence may decouple from line-level localization quality, allowing inaccurate anchors to persist before non-maximum suppression (NMS). We propose a structure-enhanced and quality-aware framework that improves lane representation and dynamic-anchor scoring while preserving the inference pipeline of the Anchor Decomposition Network (ADNet). Specifically, a Gated Horizontal-Vertical Token (GHVT) module enhances mid- and high-level backbone features via lightweight directional token interactions with a learnable residual gate. In parallel, Line-Quality-Aware Dynamic Anchor Scoring (LQAS) calibrates existing classification logits using quality supervision, hard-negative suppression, and pairwise ranking without adding inference branches. On the VIL-100 dataset, our method improves ADNet-R34 from 89.97 to 91.28 in F1 score at the 0.5 intersection-over-union threshold (F1@50), reducing both false positives and false negatives. Additional experiments on CULane and TuSimple datasets, extensive ablations, score-distribution diagnostics, and runtime analysis confirm complementary structural and ranking improvements with minimal computational overhead.
\end{abstract}

\section{Introduction}
\label{sec:introduction}

Lane detection is a core perception task for autonomous driving and advanced driver-assistance systems, estimating lane instances from forward-facing camera images to support localization, lane keeping, trajectory planning, and safety monitoring. The task remains challenging because lane markings are thin, elongated, sparse, and frequently interrupted by occlusion, poor illumination, worn paint, glare, shadows, and dense traffic. A robust detector must therefore infer coherent lane-set structure across the image, rather than relying solely on local pixel evidence or fragmented visual cues.

Recent methods advance lane detection through diverse representations and context-modeling strategies, including segmentation-based mask prediction \cite{neven2018lanenet,hou2019sad,li2023robust}, spatial-propagation and spatio-temporal aggregation \cite{pan2018scnn,zheng2021resa,dong2023hybrid,zou2019tvt,patil2026efficient}, and keypoint-, curve-, and Transformer-based modeling \cite{qu2021fololane,wang2022ganet,tabelini2020polylanenet,feng2022bezier,liu2021lstr,han2022laneformer}. These advances improve lane geometry modeling, but robust detection in difficult scenes still depends on preserving continuous lane-aware features and assigning reliable confidence to candidate lanes.

Another efficient line is anchor- and candidate-based lane detection, which formulates lane prediction as candidate generation, ranking, and refinement. Row-anchor methods use lightweight row-wise classification \cite{qin2020ufld,qin2022ufldv2}, while detector-style approaches refine instance-level candidates through attention-guided anchor pooling \cite{tabelini2021laneatt}, conditional heads \cite{liu2021condlanenet}, cross-layer priors \cite{zheng2022clrnet}, or dynamic anchor decomposition \cite{xiao2023adnet}. Despite their efficiency, these detectors remain vulnerable to two coupled failure modes: backbone features may lose continuity along long and partially occluded lanes, and the classification scores used before non-maximum suppression (NMS) may not faithfully reflect line-level localization quality. As a result, poorly localized or false candidates can receive high pre-NMS scores, outrank better-localized lanes, and persist in the final predictions.

This paper addresses these two failure modes with a structure-enhanced and quality-aware framework built upon the Anchor Decomposition Network (ADNet) \cite{xiao2023adnet}, a strong dynamic-anchor lane detector. To strengthen lane representation, we introduce the Gated Horizontal-Vertical Token (GHVT) module, a lightweight residual backbone plug-in that models directional token interactions along horizontal and vertical axes and adaptively injects the resulting context through a learnable residual gate. Because it preserves feature tensor shapes and leaves the detector head, geometry decoder, and NMS pipeline unchanged, GHVT is not tied to ADNet's dynamic-anchor formulation and can serve as a general detector-agnostic backbone enhancement. To improve candidate ranking, we further propose Line-Quality-Aware Dynamic Anchor Scoring (LQAS), which calibrates ADNet's existing classification logits so that pre-NMS scores better reflect line-level localization quality.

These two components are deliberately decoupled: GHVT acts within the feature hierarchy, particularly ResNet stages 3 and 4, to strengthen the continuity of long, thin, and occluded markings, while LQAS combines quality supervision, hard-negative suppression, and pairwise ranking on the reused classification layer. This complementary design jointly improves missed-lane recovery and false-positive suppression while preserving the original ADNet inference pipeline.

The experimental results support these claims. On the VIL-100 benchmark \cite{zhang2021vil100}, the enhanced ADNet-R34 model with GHVT and LQAS improves the reproduced ADNet-R34 baseline from 89.97 to 91.28 in F1 score at the 0.5 intersection-over-union threshold (F1@50) while reducing both false positives (FPs) and false negatives (FNs), and even outperforms the reported ADNet-R101 results in \cite{xiao2023adnet} despite using the smaller ResNet-34 backbone. Together with the lightweight design of GHVT and the branch-free inference of LQAS, this indicates that the gain stems from more effective lane-structure representation and quality-aware score calibration rather than increased parameter count. On the full 34,680-frame CULane test split \cite{pan2018scnn}, GHVT provides positive cross-detector transfer evidence when integrated into both ADNet and LaneATT \cite{tabelini2021laneatt}. Extensive ablations further show that GHVT provides the primary improvement through enhanced lane-structure representation, while LQAS reduces FPs by suppressing high-scoring hard negatives.

\section{Related Work}
\label{sec:related}

\subsection{Lane Representations: Dense Masks, Keypoints, and Curves}
\label{subsec:2.1}

Lane detection methods are commonly distinguished by how they parameterize lane instances. Segmentation-based methods formulate lane detection as semantic or instance mask prediction, using instance grouping, representation learning, self-distillation, or tailored losses \cite{neven2018lanenet,hou2019sad,li2023robust}, but often require additional grouping or post-processing to obtain lane instances. To reduce dense prediction, bottom-up keypoint methods associate local lane evidence into complete lanes \cite{qu2021fololane,wang2022ganet}, while parametric methods regress compact continuous curves, such as polynomial or B\'ezier representations, to model lane geometry directly \cite{tabelini2020polylanenet,feng2022bezier}.

\subsection{Anchor- and Candidate-based Lane Detection}
\label{subsec:2.2}

Another efficient line formulates lane detection as candidate generation, ranking, and refinement, where predefined or dynamically generated anchors are classified, scored, and regressed into final lane instances. Row-anchor methods provide a lightweight variant by discretizing lane locations along predefined image rows and solving row-wise classification \cite{qin2020ufld,qin2022ufldv2}, while instance-level candidate methods improve prediction through attention-guided anchor pooling \cite{tabelini2021laneatt}, conditional convolutional heads \cite{liu2021condlanenet}, cross-layer lane priors \cite{zheng2022clrnet}, or dynamic anchor decomposition \cite{xiao2023adnet}. These designs enable efficient inference, but their final predictions remain sensitive to feature continuity and score calibration: thin or occluded lanes require coherent lane-aware responses, while pre-NMS scores should rank well-localized lanes above ambiguous or false proposals.

\subsection{Context Modeling for Lane-structure Features}
\label{subsec:2.3}

Lane markings are sparse, elongated, and frequently interrupted, making contextual aggregation important for robust detection. Spatial message-passing and recurrent aggregation propagate fragmented lane evidence across rows or columns within a frame \cite{pan2018scnn,zheng2021resa}, and spatio-temporal methods further exploit adjacent frames under occlusion, blur, and worn markings \cite{dong2023hybrid,zou2019tvt,patil2026efficient}. Transformer-style approaches model row-, column-, or instance-level dependencies with global attention \cite{liu2021lstr,han2022laneformer}, and horizontal-vertical attention has been explored for multi-frame lane detection \cite{zhang2022mhvt}. General visual modules such as residual learning \cite{he2016resnet}, self-attention \cite{vaswani2017attention}, and channel/spatial recalibration \cite{hu2018senet,woo2018cbam} also provide useful enhancement mechanisms. These studies motivate lightweight structure-aware feature enhancement that preserves the downstream detection pipeline.

\subsection{Quality-aware Scoring and Pre-NMS Ranking}
\label{subsec:2.4}

Localization-aware confidence is crucial whenever candidates are ranked before NMS. In object detection, IoU-Net predicts localization confidence \cite{jiang2018iounet}, GFL unifies classification confidence with localization quality \cite{li2020gfl}, and VarifocalNet learns an IoU-aware classification score \cite{zhang2021varifocalnet}. Lane detection faces an analogous problem, where candidate scores should reflect line-level localization quality rather than foreground likelihood alone; CLRerNet addresses this by introducing LaneIoU into assignment and loss design \cite{honda2024clrernet}. However, quality-aware scoring for dynamic-anchor lane detectors remains underexplored, especially when the goal is to recalibrate existing classification logits without adding an inference branch or changing geometry decoding.

\section{Method}
\label{sec:method}

\subsection{Overview}
\label{subsec:overview}

\begin{figure*}[t]
   \centering
   \includegraphics[width=\textwidth]{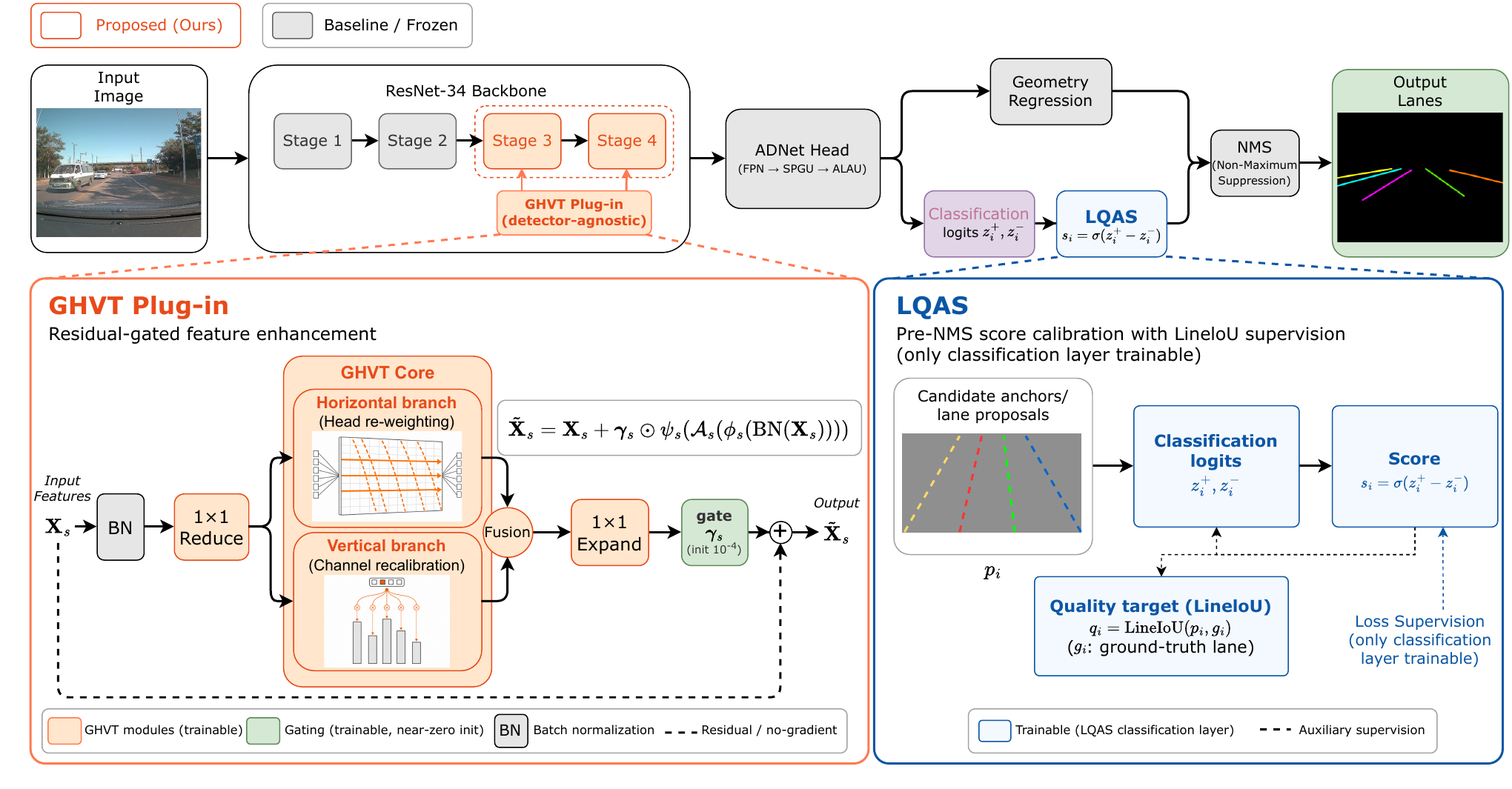}
   \caption{
   Overview of the proposed lane detection framework. 
   }
   \label{fig:method_overview}
 \end{figure*}

The proposed framework improves the dynamic-anchor lane detector ADNet \cite{xiao2023adnet} from two complementary perspectives: lane-structure representation and score calibration, while preserving the original inference pipeline. As illustrated in \Cref{fig:method_overview}, an input image is first processed by a ResNet-34 backbone, where the proposed GHVT plug-in is applied to the stage-3 and stage-4 feature maps to strengthen mid- and high-level lane structure representations. The enhanced feature hierarchy is subsequently fed into the unchanged ADNet head, which consists of a Feature Pyramid Network (FPN), a Start Point Generate Unit (SPGU), and an Adaptive Lane Aware Unit (ALAU), followed by the original lane-geometry regression and candidate classification branches. For score calibration, LQAS reuses the existing classification logits and supervises them with line-level quality targets, encouraging pre-NMS scores to better reflect localization quality without introducing an additional inference branch.

Formally, given an input image $I$, the framework can be summarized as
\begin{align}
\tilde{\mathcal F}
&= \mathcal T_{\mathrm{GHVT}}\!\left(\mathcal B(I)\right),
\label{eq:overview_feature} \\
\{(\hat{\ell}_i,s_i)\}_{i=1}^{M_{\mathrm{cand}}}
&= \mathcal D_{\mathrm{ADNet}}\!\left(\tilde{\mathcal F}\right),
\label{eq:overview_detection}
\end{align}
where $\mathcal B$ is the ResNet-34 backbone, $\mathcal T_{\mathrm{GHVT}}$ the GHVT feature-enhancement module, and $\mathcal D_{\mathrm{ADNet}}$ the unchanged ADNet detector. $M_{\mathrm{cand}}$ is the number of pre-NMS dynamic-anchor candidates, $\hat{\ell}_i$ the $i$-th lane hypothesis, and $s_i$ its pre-NMS ranking score computed from the existing classification logits. Since GHVT enhances features while LQAS modifies only score supervision, the detector head, geometry decoder, ranking rule, and NMS procedure remain unchanged during inference. During training, GHVT is optimized with the original ADNet loss, and LQAS provides an additional score-calibration loss for the classification layer, as detailed in \Cref{subsec:hvt} and \Cref{subsec:lqas}.

\subsection{GHVT for Lane-Structure Enhancement}
\label{subsec:hvt}

GHVT is a residual backbone enhancement module for strengthening elongated lane features. Given a stage feature map $\mathbf X_s \in \mathbb{R}^{C_s \times H_s \times W_s}$, GHVT applies normalization and channel compression, performs horizontal-vertical token modeling, projects the features back to the original dimension, and injects the enhanced representation through a learnable near-identity residual gate:
\begin{equation}
\tilde{\mathbf X}_s =
\mathbf X_s +
\boldsymbol{\gamma}_s \odot
\psi_s\!\left(
\mathcal A_s\!\left(\phi_s(\mathrm{BN}(\mathbf X_s))\right)
\right),
\label{eq:hvt}
\end{equation}
where $\phi_s$ and $\psi_s$ denote $1{\times}1$ channel reduction and expansion projections, BN denotes batch normalization, $\mathcal A_s$ is the horizontal-vertical token operator, $\odot$ represents element-wise multiplication, and $\boldsymbol{\gamma}_s$ is a learnable residual gate initialized to $10^{-4}$ to stabilize early-stage training by preserving the original ADNet representation.

The operator $\mathcal A_s$ captures complementary spatial and channel dependencies. Let $\mathbf T_s \in \mathbb{R}^{N_s \times d_s}$ be the flattened feature map with $N_s$ spatial tokens and embedding dimension $d_s$. The horizontal branch produces a spatial representation $\mathbf H_s$ via multi-head token attention, while the vertical branch generates a channel-wise gating vector $\mathbf w_s$ from globally aggregated context:
\begin{align}
\mathcal A_s(\mathbf T_s) &= \mathbf H_s \odot \mathbf w_s, \label{eq:hvt_fuse} \\
\mathbf H_s &= \operatorname{Concat}_{h=1}^{H} \operatorname{Attn}_h(\mathbf T_s), \label{eq:hvt_h} \\
\mathbf w_s &= \sigma\!\left(\mathbf W_2\,\delta\!\left(\mathbf W_1\,\mathrm{GAP}(\mathbf T_s)\right)\right), \label{eq:hvt_w}
\end{align}
where $\mathrm{GAP}$ denotes global average pooling, $\mathbf W_1$ and $\mathbf W_2$ form a bottleneck transformation, $\delta(\cdot)$ and $\sigma(\cdot)$ denote ReLU and sigmoid activations, $\operatorname{Concat}_{h=1}^{H}$ concatenates the $H$ attention-head outputs along the channel dimension, and $\odot$ performs element-wise channel reweighting between the spatial features $\mathbf H_s$ and channel importance $\mathbf w_s$.

In implementation, GHVT is applied to ResNet stages 3 and 4: stage 3 operates on higher-resolution features with reduced token dimensionality for efficiency, while stage 4 captures global context from low-resolution representations.

The design is detector-agnostic: since GHVT only transforms backbone feature maps while preserving their tensor shape, it is independent of ADNet's dynamic-anchor head, lane parameterization, and NMS procedure, enabling direct transfer to other lane detectors such as LaneATT \cite{tabelini2021laneatt}.

\subsection{Line-Quality-Aware Dynamic Anchor Scoring (LQAS)}
\label{subsec:lqas}

LQAS aligns ADNet \cite{xiao2023adnet} classification confidence with line-level localization quality. ADNet ranks dynamic-anchor candidates using classification confidence before NMS. However, binary foreground confidence is not always consistent with lane localization quality: poorly localized or background anchors may still receive high pre-NMS scores. LQAS addresses this mismatch by recalibrating the existing classification logits into quality-aware candidate scores, without introducing an additional inference branch or modifying the geometry decoder.

For each dynamic anchor $i$, the detector predicts a lane hypothesis $\hat{\ell}_i$ and two classification logits $(z_i^+, z_i^-)$. We use the logit margin to define the pre-NMS ranking score as:
\begin{align}
s_i = \sigma(z_i^+ - z_i^-),
\label{eq:lqas_margin}
\end{align}

For a matched anchor set $\mathcal P$ and an unmatched anchor set $\mathcal N$, LQAS assigns continuous quality targets, $q$, to positive samples based on Line Intersection-over-Union (LineIoU) \cite{honda2024clrernet}, while assigning zero to negative samples:
\begin{equation}
q_p =
\left[
\operatorname{clip}
\left(
\operatorname{LineIoU}(\hat{\ell}_p, g_{\pi(p)}; w), 0, 1
\right)
\right]^{\beta},
\quad p \in \mathcal P;\quad
q_j = 0,\; j \in \mathcal N .
\label{eq:lqas_quality}
\end{equation}

Here, $g_{\pi(p)}$ denotes the ground-truth lane assigned to anchor $p$, and LineIoU follows the lane-overlap formulation in confidence-aware lane detection \cite{honda2024clrernet}. The parameter $w$ controls the spatial tolerance in LineIoU computation, while $\beta$ is a fixed calibration exponent controlling the sharpness of the quality target. This formulation encourages classification scores to reflect line-level localization quality rather than merely foreground likelihood.

To focus supervision on candidates that influence NMS, we mine score-adaptive hard negatives from unmatched anchors $\mathcal N$:
\begin{equation}
\mathcal H =
\operatorname*{Top\text{-}K}_{j \in \mathcal N}(s_j, K), \quad
K = \min\left(|\mathcal N|, \max(K_{\min}, \lfloor r|\mathcal P| \rfloor)\right).
\label{eq:lqas_hard}
\end{equation}
Here, $\mathcal H$ denotes the set of selected hard-negative anchors, and $s_j$ is the pre-NMS ranking score of candidate $j \in \mathcal N$. The parameter $K$ controls the number of mined hard negatives, where $K_{\min}$ ensures a minimum selection size and $r$ defines the proportional sampling ratio with respect to the positive set $\mathcal P$.

The final LQAS loss combines quality alignment, hard-negative suppression, and pairwise ranking:
\begin{equation}
\begin{aligned}
\mathcal L_{\mathrm{LQAS}} &=
\frac{1}{|\mathcal P|}
\sum_{p \in \mathcal P}
\mathrm{BCE}(s_p, q_p)
+
\lambda_{\mathrm h}
\frac{1}{|\mathcal H|}
\sum_{h \in \mathcal H}
\mathrm{BCE}(s_h, 0) \\
&\quad+
\lambda_{\mathrm r}
\frac{1}{|\mathcal P||\mathcal H|}
\sum_{p \in \mathcal P}
\sum_{h \in \mathcal H}
\left[\tau + s_h - s_p\right]_+ .
\end{aligned}
\label{eq:lqas_loss}
\end{equation}

Here, $\mathcal P$ and $\mathcal H$ denote the sets of positive and hard-negative anchors, respectively. $s_p$ is the pre-NMS ranking score, and $q_p$ is the LineIoU-based quality target. $\mathrm{BCE}(\cdot,\cdot)$ denotes the binary cross-entropy loss, while $[\cdot]_+ = \max(\cdot, 0)$ denotes the hinge function with margin $\tau$. The hyperparameters $\lambda_{\mathrm h}$ and $\lambda_{\mathrm r}$ balance hard-negative suppression and ranking supervision, respectively.

Overall, the first term in \Cref{eq:lqas_loss} enforces alignment between scores and continuous lane quality, the second term suppresses high-scoring hard negatives, and the third term explicitly encourages correct ranking of positives over confusing negatives before NMS.

In implementation, LQAS is trained on top of the GHVT-enhanced ADNet checkpoint with a narrow trainable scope: only the classification layer is updated, while the backbone, GHVT, FPN, SPGU, ALAU, and regression head are frozen. This design isolates score-quality alignment from geometry learning and ensures that the observed improvements are attributed to enhanced pre-NMS ranking rather than additional inference complexity.

\subsection{Inference}
\label{subsec:inference}
The proposed model preserves the original ADNet inference pipeline. At test time, dynamic-anchor candidates are decoded and ranked using the same score formulation and logic described in \Cref{subsec:lqas}, followed by the unchanged NMS and lane decoding procedure. No additional inference branches are introduced, and the computational complexity remains identical to ADNet apart from the GHVT-enhanced backbone.

\section{Experiments and Results}
\label{sec:experiments}

\subsection{Experimental Setup}
\label{subsec:setup}

We evaluate the proposed method on three lane-detection benchmarks with complementary scene characteristics: VIL-100 \cite{zhang2021vil100}, CULane \cite{pan2018scnn}, and TuSimple \cite{tusimple}. VIL-100 \cite{zhang2021vil100} serves as the primary benchmark, as it contains video-derived driving scenes with frequent occlusion, illumination variation, dense lane layouts, and challenging road structures. CULane \cite{pan2018scnn} provides diverse real-world scenarios and challenging difficult subsets, including crowded, night, and no-line conditions; we use its full 34,680-frame test split to evaluate cross-dataset generalization performance beyond VIL-100. TuSimple \cite{tusimple} consists of relatively structured highway scenes and is used as an auxiliary benchmark under simpler geometric conditions.

For all datasets, we report standard lane-detection metrics, including F1@50, accuracy (Acc), false positive (FP) rate, and false negative (FN) rate, and additionally report F1@75 on CULane to evaluate performance under stricter localization requirements. All local training, fine-tuning, testing, and runtime measurements are conducted on a single NVIDIA A40 GPU with CUDA 12.1 and 128 GB system RAM to ensure fair comparisons.

\subsection{Main Results on the VIL-100 Benchmark}
\label{subsec:vil100}

\begin{table}[!htbp]
\caption{Quantitative comparison of lane detection methods on VIL-100 benchmark.}
\label{tab:vil100_main}
\centering
\small
\setlength{\tabcolsep}{4pt}
\begin{tabular*}{\textwidth}{@{\extracolsep{\fill}}lrrrr@{}}
\toprule
Method & F1@50↑ & Acc (\%)↑ & FP (\%)↓ & FN (\%)↓ \\
\midrule
MMA-Net \cite{zhang2021vil100} & 83.90 & 91.00 & 11.10 & 10.50 \\
MHVT \cite{zhang2022mhvt} & 84.60 & 91.50 & 8.20 & 9.50 \\
LaneNet \cite{neven2018lanenet} & 72.10 & 85.80 & 12.20 & 20.70 \\
SCNN-VGG16 \cite{pan2018scnn} & 49.10 & 90.70 & 12.80 & 11.00 \\
SAD-ENet \cite{hou2019sad} & 75.50 & 88.60 & 17.00 & 15.20 \\
UFLD-R34 \cite{qin2020ufld} & 31.00 & 85.20 & 11.50 & 21.50 \\
LSTR \cite{liu2021lstr} & 70.30 & 88.40 & 16.30 & 14.80 \\
CLRNet-R18 \cite{zheng2022clrnet} & 57.27 & 88.99 & 6.90 & 13.50 \\
CLRNet-R101 \cite{zheng2022clrnet} & 59.41 & 88.65 & 2.10 & 12.50 \\
ADNet-R34 \cite{xiao2023adnet} & 90.39 & 94.38 & 4.40 & 4.90 \\
ADNet-R101 \cite{xiao2023adnet} & 90.90 & 94.27 & 4.70 & 5.00 \\
\midrule
Baseline ADNet-R34 & 89.97 & 94.38 & 4.43 & 4.90 \\
\textbf{Enhanced ADNet-R34} & \textbf{91.28} & \textbf{94.43} & \textbf{4.26} & \textbf{4.69} \\
\bottomrule
\end{tabular*}
\end{table}

The proposed framework achieves strong performance on the VIL-100 benchmark. As shown in \cref{tab:vil100_main}, the enhanced ADNet-R34 with the proposed GHVT and LQAS (ADNet-R34\_GHVT+LQAS) achieves the F1@50 of 91.28, surpassing the reproduced ADNet-R34 baseline by 1.31 points. This improvement is accompanied by consistent reductions in both FP (4.43\% $\rightarrow$ 4.26\%) and FN (4.90\% $\rightarrow$ 4.69\%), indicating that the gain is not obtained through a precision–recall trade-off. The proposed method also surpasses the ADNet-R101 results in \cite{xiao2023adnet}, which uses a larger backbone, across all evaluated metrics. This suggests the performance gain is not due to increased model complexity or parameter count, but rather to more effective lane-structure modeling and quality-aware scoring.

\begin{figure}[!htbp]
\centering
\includegraphics[width=0.905\linewidth]{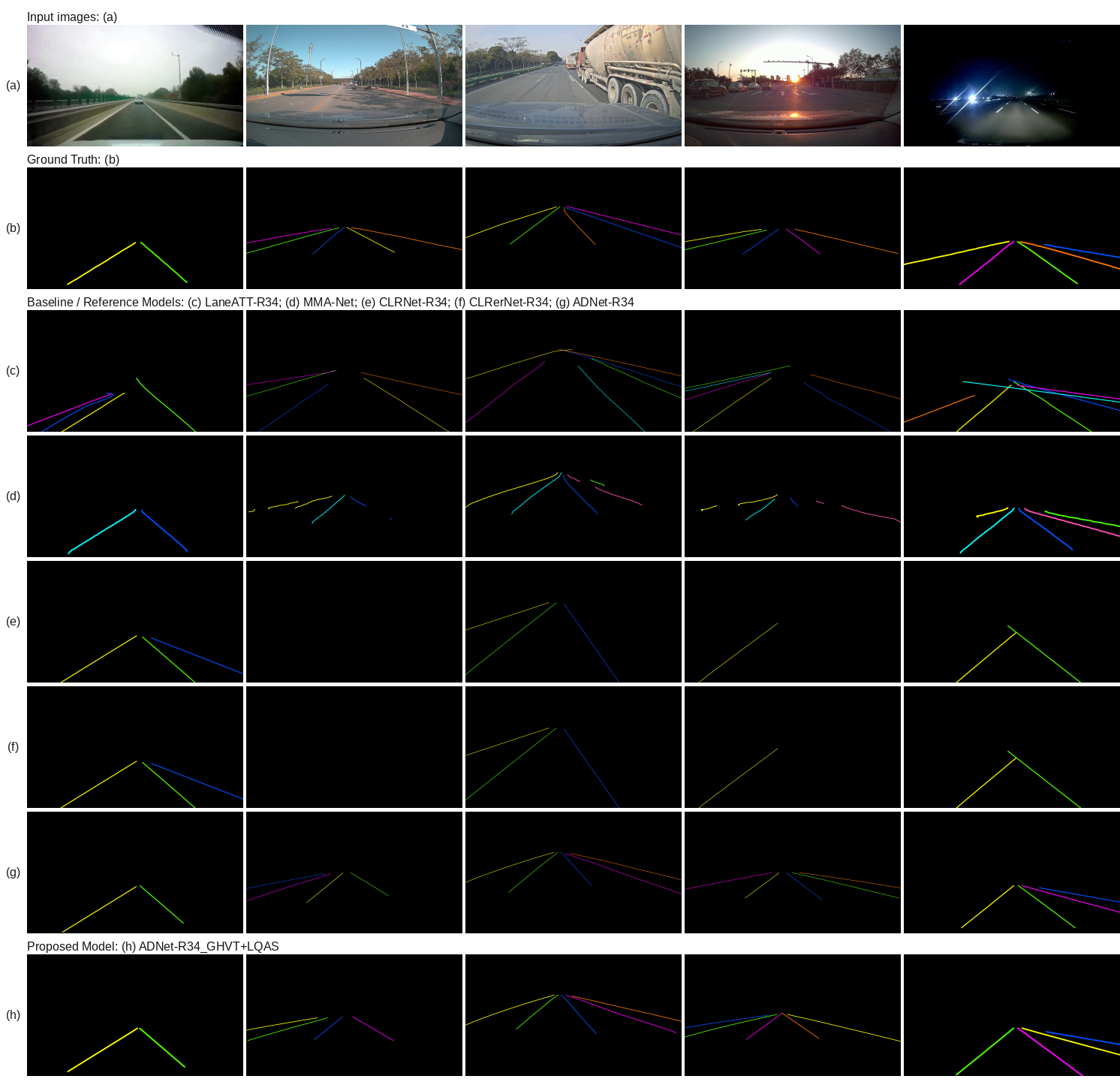}
\caption{Qualitative comparison on the VIL-100 benchmark. The selected scenes include fog, dense lane layout, truck occlusion, glare, and night driving.}
\label{fig:vil100_qual}
\end{figure}

Consistent with these quantitative results, the qualitative comparisons in \cref{fig:vil100_qual} show that the enhanced ADNet-R34\_GHVT+LQAS produces more complete and stable lane hypotheses under challenging conditions, including fog, glare, night driving, dense traffic, and heavy occlusion. In these scenarios, the baseline model is more prone to fragmented or distracted predictions due to incomplete markings and surrounding structural interference.

\subsection{Cross-Dataset Generalization and Transfer Evidence}
\label{subsec:transfer}

This section evaluates whether the proposed representation enhancement remains effective beyond the primary VIL-100 benchmark.

\begin{figure}[!htbp]
\centering
\includegraphics[width=0.912\linewidth]{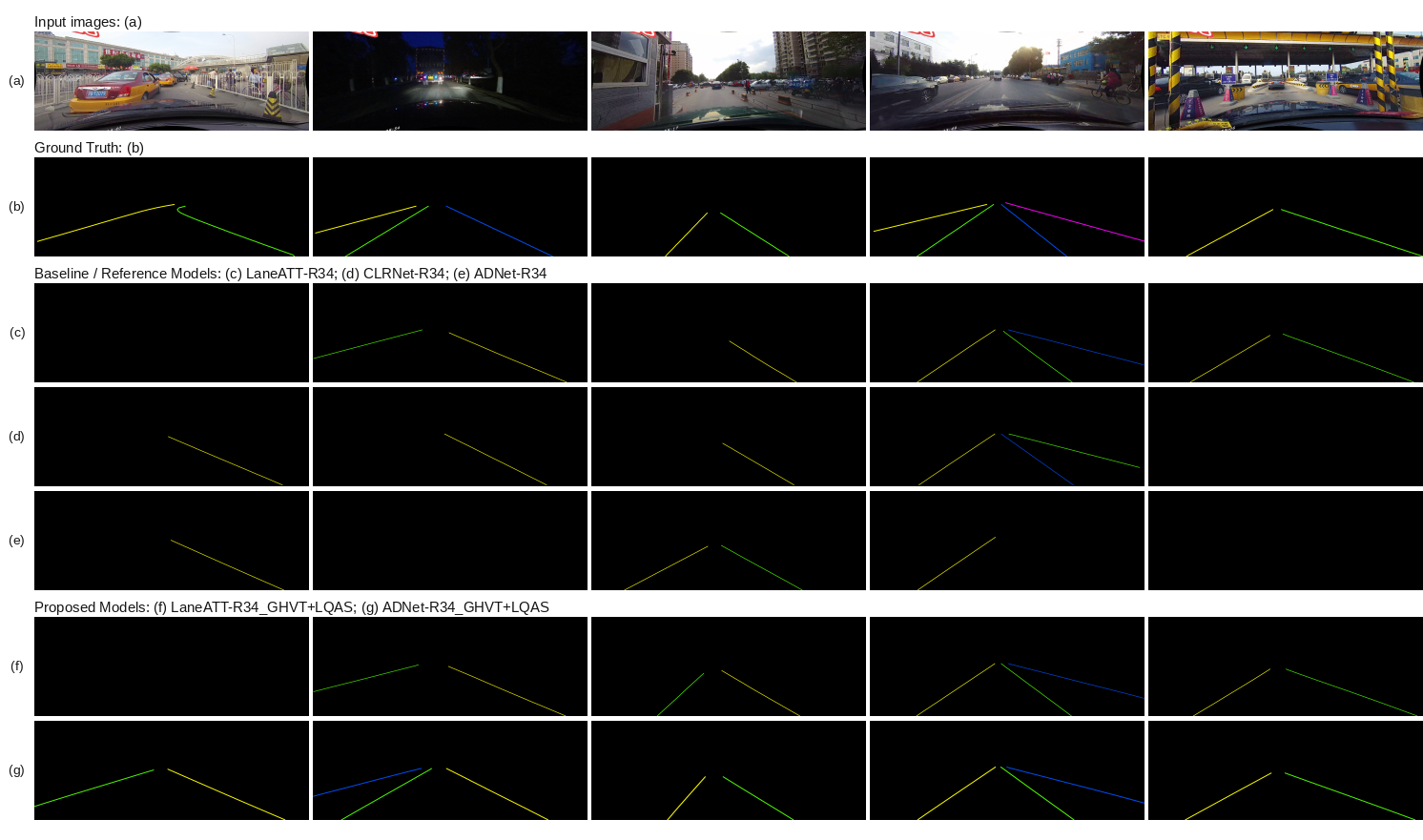}
\caption{Qualitative comparison on CULane benchmark in crowd, night, normal, arrow-marking, and no-line scenarios.}
\label{fig:culane_qual}
\end{figure}

\paragraph{Results on CULane.}
\Cref{tab:culane_main} reports both literature and locally reproduced results on the full 34,680-frame CULane test split. The enhanced ADNet-R34\_GHVT+LQAS improves the ADNet-R34 baseline from 78.95 to 79.09 in F1@50 and from 60.65 to 60.86 in F1@75. Similarly, the enhanced LaneATT-R34 with GHVT and LQAS (LaneATT-R34\_GHVT+LQAS) improves the corresponding baseline from 76.58 to 76.75 in F1@50 and from 53.62 to 54.63 in F1@75. These consistent improvements across two different detector families indicate that the proposed GHVT and LQAS generalize effectively beyond ADNet backbone and the VIL-100 training distribution. More importantly, the gains are observed across multiple challenging subsets, including crowded, night, and no-line scenarios, as shown in \Cref{tab:culane_main}. The qualitative results in \Cref{fig:culane_qual} further corroborate this observation with ADNet-R34\_GHVT+LQAS and LaneATT-R34\_GHVT+LQAS outperforming their corresponding baseline by more correctly detected lane line numbers and more stable lane lines.

% Put this macro before the table, or in the preamble.
\providecommand{\gaincell}[2]{%
  \begin{tabular}[c]{@{}c@{}}#1\\[-1pt]{\tiny $(#2)$}\end{tabular}%
}

% Put this macro before the table, or in the preamble.
\providecommand{\gaincell}[2]{%
  \begin{tabular}[c]{@{}c@{}}#1\\[-2pt]{\fontsize{5.8pt}{5.8pt}\selectfont $(#2)$}\end{tabular}%
}

\begin{table}[!htbp]
\caption{CULane benchmark evaluation. Literature results are reported for reference, while reproduced results assess the cross-dataset and cross-detector generalization of the proposed GHVT and LQAS. Values in parentheses indicate gains/decreases over the corresponding baseline.}
\label{tab:culane_main}
\centering
\small%\scriptsize
\setlength{\tabcolsep}{4pt}
\renewcommand{\arraystretch}{1.08}
\resizebox{\textwidth}{!}{%
\begin{tabular}{@{}lcccccccc@{}}
\toprule
Method & Backbone & F1@50$\uparrow$ & F1@75$\uparrow$ & Normal$\uparrow$ & Crowd$\uparrow$ & NoLine$\uparrow$ & Cross$\downarrow$ & Night$\uparrow$ \\
\midrule
SCNN \cite{pan2018scnn} & VGG16 & 71.60 & N/R & 90.60 & 69.70 & 43.40 & 1990 & 66.10 \\
LaneAF \cite{abualsaud2021laneaf} & DLA34 & 77.41 & N/R & 91.80 & 75.61 & 51.38 & 1360 & 73.03 \\
GANet \cite{wang2022ganet} & R34 & 79.39 & N/R & 93.73 & 77.92 & 52.63 & 1368 & 73.67 \\
CondLane \cite{liu2021condlanenet} & R34 & 78.74 & 59.39 & 93.38 & 77.14 & 51.85 & 1387 & 73.92 \\
CLRNet \cite{zheng2022clrnet} & R34 & 79.73 & 62.11 & 93.49 & 78.06 & 54.01 & 1216 & 75.02 \\
CLRerNet \cite{honda2024clrernet} & R34 & 80.76 & 63.77 & 93.93 & 79.51 & 55.55 & 1088 & 76.02 \\
ADNet \cite{xiao2023adnet} & R34 & 78.94 & N/R & 92.90 & 77.45 & 52.89 & 1499 & 74.78 \\
\midrule
Baseline ADNet-R34 & R34 & 78.95 & 60.65 & 92.90 & 77.46 & 52.89 & 1498 & 74.78 \\
Enhanced ADNet-R34 & R34
& \gaincell{79.09}{+0.14}
& \gaincell{60.86}{+0.21}
& \gaincell{93.46}{+0.56}
& \gaincell{77.58}{+0.12}
& \gaincell{53.49}{+0.60}
& \gaincell{1499}{-1}
& \gaincell{74.99}{+0.21} \\
Baseline LaneATT-R34  & R34 & 76.58 & 53.62 & 92.12 & 74.84 & 49.22 & 1310 & 70.54 \\
Enhanced LaneATT-R34 & R34
& \gaincell{76.75}{+0.17}
& \gaincell{54.63}{+1.01}
& \gaincell{92.37}{+0.25}
& \gaincell{74.79}{-0.05}
& \gaincell{49.98}{+0.76}
& \gaincell{1311}{-1}
& \gaincell{71.18}{+0.64} \\
\bottomrule
\end{tabular}%
}
\end{table}

\paragraph{Results on TuSimple.} The TuSimple benchmark provides auxiliary evidence under a more structured highway setting. As shown in \Cref{tab:tusimple_main}, the reproduced ADNet-R34 baseline matches the reported ADNet-R34 results in \cite{xiao2023adnet}. Incorporating the proposed GHVT and LQAS modules improves performance, increasing F1@50 from 97.31 to 97.49 while reducing the FP rate from 2.83\% to 2.41\%. These results indicate that the proposed model enhancement provides precision-oriented transfer gains under the simpler lane geometry in TuSimple.

\subsection{Ablation Study of GHVT and LQAS on VIL-100}
\label{subsec:component_ablation}

We conduct a sequential ablation on the VIL-100 benchmark to isolate the contributions of the two proposed components. Starting from ADNet-R34 baseline, we first add the GHVT module to evaluate lane-structure representation and then add the LQAS module to evaluate quality-aware score calibration for pre-NMS candidate ranking.

\FloatBarrier

\begin{table}[!htbp]
\caption{TuSimple benchmark evaluation. }
\label{tab:tusimple_main}
\centering
\small
\setlength{\tabcolsep}{3pt}
\renewcommand{\arraystretch}{0.9}

\begin{tabular*}{\textwidth}{@{\extracolsep{\fill}}llrrrr@{}}
\toprule
Method & Backbone & F1@50↑ & Acc(\%)↑ & FP(\%)↓ & FN(\%)↓ \\
\midrule
UFLDv2 \cite{qin2022ufldv2} & R34 & 96.22 & 95.56 & 3.18 & 4.37 \\
LaneATT \cite{tabelini2021laneatt} & R34 & 96.77 & 95.63 & 3.53 & 2.92 \\
CondLaneNet \cite{liu2021condlanenet} & R101 & 97.24 & 96.54 & 2.01 & 3.50 \\
ADNet-R34 \cite{xiao2023adnet} & R34 & 97.31 & 96.60 & 2.83 & 2.53 \\
\midrule
Baseline ADNet-R34 & R34 & 97.31 & 96.60 & 2.83 & 2.53 \\
Enhanced ADNet-R34 & R34 & \textbf{97.49} & 96.49 & 2.41 & 2.62 \\
\bottomrule
\end{tabular*}
\end{table}

As shown in \Cref{tab:component_ablation}, adding GHVT provides the dominant gain, improving F1@50 from 89.97 to 91.21 (+1.24), while increasing the number of true positives (TPs) from 7580 to 7702 and reducing FNs from 869 to 747. This indicates improved lane recovery. Adding LQAS further improves F1@50 to 91.28 and yields the best precision, mainly by reducing FPs from 738 to 712 while keeping recall rate close to ADNet\_GHVT. These results support the intended complementarity: GHVT improves structural lane recovery, whereas LQAS improves candidate ranking and false-positive suppression before NMS.

\begin{table}[!htbp]
\caption{Sequential component ablation of GHVT and LQAS on VIL-100 benchmark.}
\label{tab:component_ablation}
\centering
\small
\setlength{\tabcolsep}{3pt}
\renewcommand{\arraystretch}{0.9}

\begin{tabular*}{\textwidth}{@{\extracolsep{\fill}}lccccccc@{}}
\toprule
Model & TPs↑ & FPs↓ & FNs↓ & Prec.(\%)↑ & Rec.(\%)↑ & F1@50↑ & $\Delta$F1@50↑ \\
\midrule
ADNet-R34 (baseline) & 7580 & 821 & 869 & 90.23 & 89.71 & 89.97 & 0.00 \\
ADNet\_GHVT & 7702 & 738 & 747 & 91.26 & 91.16 & 91.21 & +1.24 \\
\makecell[l]{{ADNet\_GHVT+LQAS}} & 7692 & \textbf{712} & 757 & \textbf{91.53} & 91.04 & \textbf{91.28} & \textbf{+1.31} \\
\bottomrule
\end{tabular*}
\end{table}

\subsection{Analysis of GHVT for Lane-Structure Enhancement}
\label{subsec:ghvt_analysis}

\paragraph{Branch and stage placement.}
\Cref{tab:hvt_stage} analyzes both the branch composition and insertion stage of GHVT. For branch composition, the branch-removal variants keep the module inserted at stages 3 and 4, but remove either the horizontal or vertical branch. The results show that GHVT is not a generic attention add-on: using only the horizontal branch leads to severe degradation, whereas the vertical-only variant still improves model performance over the ADNet-R34 baseline. This suggests that the vertical channel-recalibration branch is important for stabilizing the enhanced representation, and that horizontal token interactions become effective when coupled with channel-wise reliability modeling. For stage placement, stage 3 contributes more than stage 4, indicating that lane detection benefits from enhancing features while thin lane evidence remains spatially resolvable. Combining stages 3 and 4 with full GHVT branches achieves the best result among those variants, suggesting complementarity between mid-level spatial detail and high-level context.

\begin{table}[!htbp]

\caption{GHVT branch and stage-placement ablation on the VIL-100 benchmark. Rows 2--3 remove either the vertical or horizontal branch with GHVT inserted at stages 3+4, while rows 4--5 evaluate the full GHVT block at individual backbone stages.}
\label{tab:hvt_stage}
\centering
\small
\setlength{\tabcolsep}{3pt}
\renewcommand{\arraystretch}{0.918}
\begin{tabular*}{\textwidth}{@{}>{\raggedright\arraybackslash}p{0.28\textwidth}>{\raggedright\arraybackslash}p{0.42\textwidth}@{\extracolsep{\fill}}rr@{}}
\toprule
Variant & Placement & F1@50↑ & $\Delta$F1@50↑ \\
\midrule
ADNet-R34 & none & 89.97 & 0.00 \\
GHVT w/o vertical & stages 3+4, horizontal only & 77.31 & -12.66 \\
GHVT w/o horizontal & stages 3+4, vertical only & 90.77 & +0.79 \\
Stage 4 only & stage 4 & 90.79 & +0.82 \\
Stage 3 only & stage 3 & 91.11 & +1.14 \\
Full GHVT & stages 3+4 & \textbf{91.21} & \textbf{+1.24} \\
\bottomrule
\end{tabular*}
\end{table}

\paragraph{Difficult-case analysis and feature visualization.}
We next examine whether GHVT improves the performance in scenes where lane-structure modeling is most critical. \Cref{tab:hvt_subset} reports subset results on VIL-100. Sample sizes in each subset are counted from annotations, while F1@50, FPs, and FNs follow the same validation protocol as the main evaluation. The ADNet-R34 and ADNet-R34\_GHVT columns report F1@50, and $\Delta$F1@50, $\Delta$FPs, and $\Delta$FNs are computed as the performance of ADNet-R34\_GHVT minus that of ADNet-R34.

% \begin{center}
% \begin{minipage}{\textwidth}
% \captionof{table}{GHVT performance analysis across different subsets on VIL-100 benchmark.}
% \label{tab:hvt_subset}
% \vspace{3pt}

% \scriptsize
% \setlength{\tabcolsep}{1.6pt}
% \begin{tabular*}{\textwidth}{@{}>{\raggedright\arraybackslash}p{0.25\textwidth}@{\extracolsep{\fill}}rrrrrrr@{}}
% \toprule
% Subset & \#Img & \#Lane & ADNet-R34 & ADNet-R34\_GHVT & $\Delta$F1@50 & $\Delta$FPs & $\Delta$FNs \\
% \midrule
% All test set & 2000 & 8451 & 89.97 & 91.21 & +1.24 & -83 & -122 \\
% Non-occluded & 244 & 934 & 92.57 & 92.41 & -0.16 & +1 & +2 \\
% Partial occlusion & 279 & 1252 & 85.40 & 87.08 & +1.68 & -11 & -28 \\
% Severe partial occlusion & 858 & 3815 & 88.51 & 89.79 & +1.28 & -51 & -47 \\
% Fully occlusion & 619 & 2450 & 93.53 & 95.00 & +1.48 & -22 & -49 \\
% Four-lane scenes & 835 & 3341 & 90.23 & 91.78 & +1.55 & -51 & -54 \\
% Five-lane scenes & 911 & 4556 & 88.94 & 89.97 & +1.03 & -27 & -62 \\
% Occluded multi-lane & 1546 & 6987 & 88.97 & 90.41 & +1.44 & -78 & -118 \\
% \bottomrule
% \end{tabular*}
% \end{minipage}
% \end{center}

\begin{table}[!htbp]
\caption{GHVT performance analysis across different subsets on VIL-100 benchmark.}
\label{tab:hvt_subset}
\centering
%\vspace{2pt}
%\footnotesize

\scriptsize
\setlength{\tabcolsep}{2.6pt}
\begin{tabular*}{\textwidth}{@{}>{\raggedright\arraybackslash}p{0.25\textwidth}@{\extracolsep{\fill}}rrrrrrr@{}}
\toprule
Subset & \#Img & \#Lane & ADNet-R34 & \makecell[c]{ADNet-R34\_GHVT} & $\Delta$F1@50 & $\Delta$FPs & $\Delta$FNs \\
\midrule
All test set & 2000 & 8451 & 89.97 & 91.21 & +1.24 & -83 & -122 \\
Non-occluded & 244 & 934 & 92.57 & 92.41 & -0.16 & +1 & +2 \\
Partial occlusion & 279 & 1252 & 85.40 & 87.08 & +1.68 & -11 & -28 \\
Severe partial occlusion & 858 & 3815 & 88.51 & 89.79 & +1.28 & -51 & -47 \\
Full occlusion & 619 & 2450 & 93.53 & 95.00 & +1.48 & -22 & -49 \\
Four-lane scenes & 835 & 3341 & 90.23 & 91.78 & +1.55 & -51 & -54 \\
Five-lane scenes & 911 & 4556 & 88.94 & 89.97 & +1.03 & -27 & -62 \\
Occluded multi-lane & 1546 & 6987 & 88.97 & 90.41 & +1.44 & -78 & -118 \\
\bottomrule
\end{tabular*}
\end{table}

The subset results indicate that GHVT strengthens lane-structure representation, with the main gains appearing in scenes where lane continuity is difficult to recover. On the full VIL-100 test set, GHVT reduces both FPs and FNs. The improvements are especially clear in occlusion-related subsets: F1@50 increases by 1.68, 1.28, and 1.48 points for partial, severe partial, and full occlusion subsets, respectively, with consistent reductions in FPs and FNs. A similar trend is observed in multi-lane scenes, where the occluded multi-lane subset achieves a 1.44-point F1@50 gain while substantially reducing both FPs and FNs. By contrast, the non-occluded subset remains nearly unchanged. These results suggest that GHVT mainly benefits cases requiring continuous lane evidence to be recovered from occlusion, dense lane layouts, or visually incomplete cues.

\begin{figure}[!htbp]
\centering
\includegraphics[width=\linewidth]{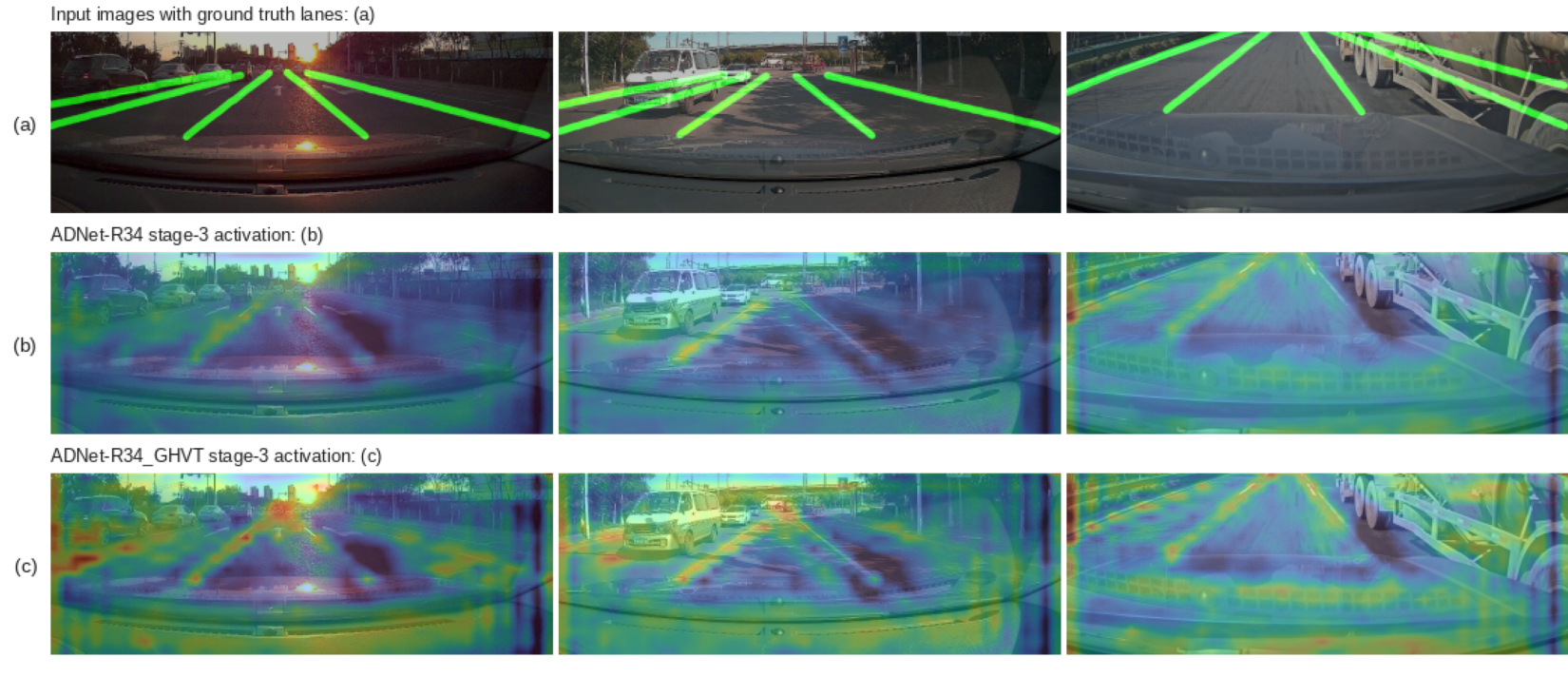}
\caption{Effect of GHVT on stage-3 feature activations. From top to bottom: (a) input images with ground-truth lanes, (b) ADNet-R34 stage-3 activations, and (c) ADNet-R34\_GHVT stage-3 activations.}
\label{fig:hvt_mechanism}
\end{figure}

\Cref{fig:hvt_mechanism} provides visual support and further illustrates the effect of GHVT. Compared with ADNet-R34, ADNet-R34\_GHVT produces more continuous stage-3 responses along lane regions and suppresses activations around vehicle occlusions, glare, road boundaries, and other distracting structures. This supports the role of GHVT in strengthening lane-structure representation.

\subsection{Analysis of LQAS for Quality-Aware Candidate Ranking}

\paragraph{Detection-level effect.}
LQAS targets residual ranking errors after GHVT enhances lane-structure representation. As shown in \Cref{tab:component_ablation}, adding LQAS to ADNet-R34\_GHVT reduces FPs from 738 to 712 and improves precision from 91.26\% to 91.53\%, while maintaining similar recall rate. This indicates that LQAS improves pre-NMS candidate ranking and false-positive suppression.

\begin{figure}[!htbp]
\centering
\includegraphics[width=0.7826\linewidth]{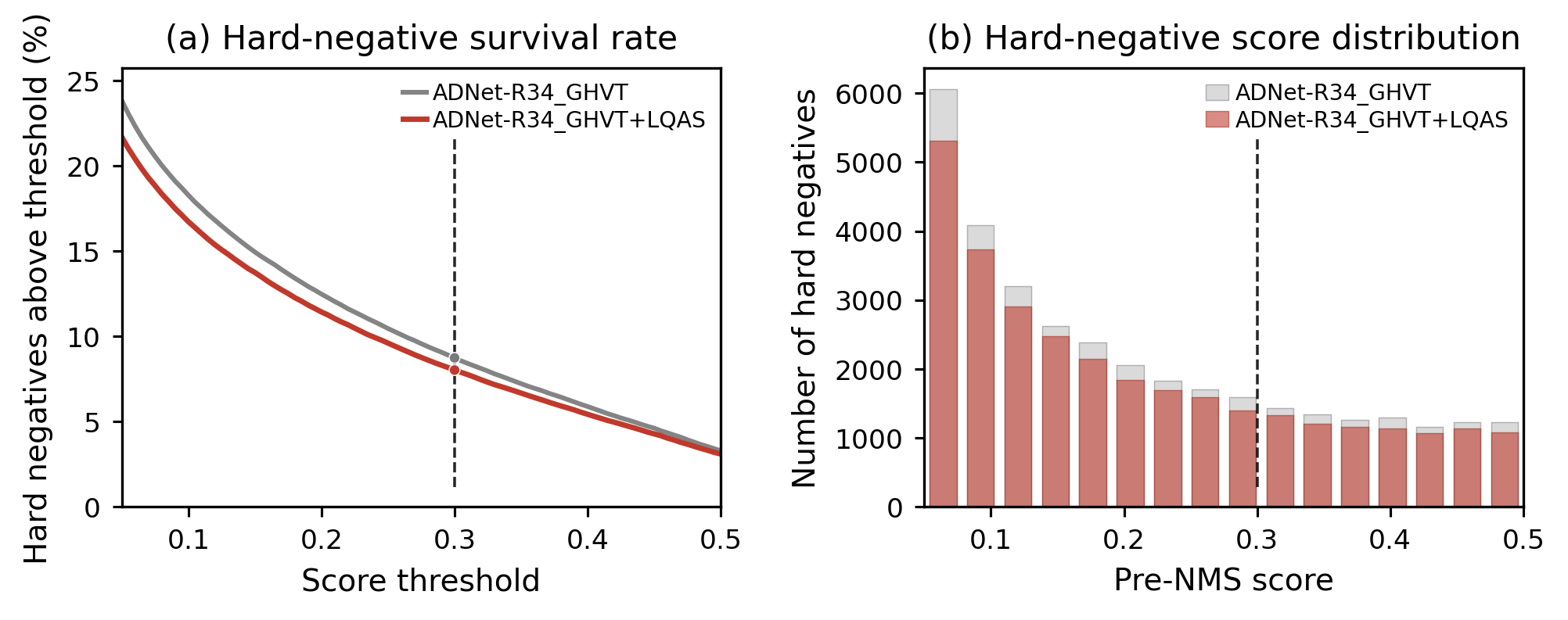}
\caption{Effect of LQAS for quality-aware pre-NMS candidate ranking on VIL-100 benchmark.}
\label{fig:lqas_scores}
\end{figure}

\paragraph{Score-distribution diagnostics.}
The score diagnostics in \Cref{tab:lqas_analysis} and \Cref{fig:lqas_scores} help explain why LQAS reduces FPs. \Cref{fig:lqas_scores}(a) reports the hard-negative survivor rate, defined as the percentage of unmatched dynamic anchors whose pre-NMS score exceeds a given threshold. \Cref{fig:lqas_scores}(b) shows the corresponding raw score distribution, where the dashed line marks the 0.30 score threshold. With LQAS, the hard-negative 90th-percentile score decreases from 0.262 to 0.238, and the proportion of hard negatives above 0.30 drops from 8.73\% to 8.02\%. Meanwhile, positive scores are preserved. These results indicate that LQAS improves pre-NMS ranking by suppressing confusing high-scoring negatives without weakening positive candidates.

%LQAS reduces the hard-negative mean score from 0.067 to 0.062 and the hard-negative 90th-percentile score from 0.262 to 0.238. It also decreases the proportion of hard negatives above the 0.30 threshold from 8.73\% to 8.02\%, while slightly increasing the positive mean score from 0.624 to 0.632. These trends indicate that LQAS improves pre-NMS ranking by suppressing confusing high-scoring negatives while preserving positive candidates before NMS.

\begin{table}[!htbp]
\caption{LQAS score-distribution diagnostics on the VIL-100 benchmark. Relative benefits are computed according to the metric direction. Lower hard-negative scores and larger positive--negative score separation indicate better pre-NMS ranking.}
\label{tab:lqas_analysis}
\centering
%\scriptsize
\footnotesize
%\fontsize{7.5pt}{8.6pt}\selectfont

\setlength{\tabcolsep}{1pt}
\renewcommand{\arraystretch}{0.9}

\begin{tabular*}{\columnwidth}{@{}l@{\extracolsep{\fill}}ccc@{}}
\toprule
Metric 
& \makecell[c]{ADNet-R34\\+GHVT} 
& \makecell[c]{ADNet-R34\\+GHVT+LQAS} 
& \makecell[c]{Relative \\benefit} \\
\midrule
Hard-neg. mean score$\downarrow$ 
& 0.067 & 0.062 & 7.46\% \\
Hard-neg. P90 score$\downarrow$ 
& 0.262 & 0.238 & 9.16\% \\
Hard-neg. survivor ($s \geq 0.30$)$\downarrow$ 
& 8.73\% & 8.02\% & 8.13\% \\
Positive mean score$\uparrow$ 
& 0.624 & 0.632 & 1.28\% \\
Pos. vs. hard-neg. separation$\uparrow$ 
& 0.556 & 0.571 & 2.70\% \\
\bottomrule
\end{tabular*}
\end{table}

\vspace{6pt}

\FloatBarrier

\subsection{Analysis of Computational Efficiency and Model Complexity}
\label{subsec:efficiency}
We evaluate whether the proposed modules introduce practical deployment overhead. \Cref{tab:efficiency} reports model size and local throughput measured on a single NVIDIA A40, with $\Delta$ values computed relative to the corresponding baselines. Since LQAS recalibrates the existing classification logits and introduces no additional inference branch, the inference-time overhead mainly comes from GHVT. The enhanced ADNet-R34\_GHVT+LQAS increases the parameter count by 1.32\% and reduces FPS by 3.31\%, while LaneATT-R34\_GHVT+LQAS increases the parameter count by 1.31\% and reduces FPS by 2.16\%. Overall, the proposed model enhancement adds only about 1.3\% parameters and preserves high inference throughput, indicating limited computational overhead.

\begin{table}[!htbp]
\caption{Runtime and model complexity measured locally on an NVIDIA A40.}
\label{tab:efficiency}
\centering
\footnotesize
%\scriptsize

%\small
\setlength{\tabcolsep}{4pt}
\renewcommand{\arraystretch}{0.9}

\begin{tabular}{@{}lcccc@{}}
\toprule
Model & Params (M) & $\Delta$Params & FPS & $\Delta$FPS \\
\midrule
ADNet-R34 & 21.95 & -- & 272 & -- \\
ADNet-R34\_GHVT+LQAS & 22.24 & +1.32\% & 263 & -3.31\% \\
LaneATT-R34 & 22.13 & -- & 231 & -- \\
LaneATT-R34\_GHVT+LQAS & 22.42 & +1.31\% & 226 & -2.16\% \\
\bottomrule
\end{tabular}
\end{table}

\section{Conclusion}
\label{sec:conclusion}

In this paper, we present a structure-enhanced and quality-aware framework for robust lane detection. The proposed GHVT module strengthens mid- and high-level backbone features via lightweight residual horizontal-vertical token modeling, while LQAS aligns ADNet’s classification confidence with line-level localization quality without introducing additional inference-time branches. On the VIL-100 benchmark, the proposed enhanced ADNet-R34 model (ADNet-R34\_GHVT+LQAS) improves the reproduced baseline from 89.97 to 91.28 in F1@50 and reduces both false positives and false negatives. Extensive evaluations, including ablation studies, difficult-subset analysis, score-distribution diagnostics, cross-dataset performance generalization on CULane and TUSimple, and runtime measurements, further demonstrate that GHVT provides the primary structural improvement, while LQAS enhances pre-NMS candidate ranking with minimal computational overhead.

% ---- Bibliography ----
\bibliographystyle{splncs04}
\bibliography{main}

@article{li2023robust,
  author  = {Li, Ruohan and Dong, Yongqi},
  title   = {Robust Lane Detection Through Self-Pretraining With Masked Sequential Autoencoders and Fine-Tuning With Customized {PolyLoss}},
  journal = {IEEE Transactions on Intelligent Transportation Systems},
  volume  = {24},
  number  = {12},
  pages   = {14121--14132},
  year    = {2023},
  doi     = {10.1109/TITS.2023.3305015}
}

@article{dong2023hybrid,
  author  = {Dong, Yongqi and Patil, Sandeep and van Arem, Bart and Farah, Haneen},
  title   = {A Hybrid Spatial--Temporal Deep Learning Architecture for Lane Detection},
  journal = {Computer-Aided Civil and Infrastructure Engineering},
  volume  = {38},
  number  = {1},
  pages   = {67--86},
  year    = {2023},
  doi     = {10.1111/mice.12829}
}

@misc{patil2026efficient,
  title         = {Efficient Sequential Neural Network with Spatial-Temporal Attention and Linear {LSTM} for Robust Lane Detection Using Multi-Frame Images},
  author        = {Patil, Sandeep and Dong, Yongqi and Farah, Haneen and Hellendoorn, Hans},
  year          = {2026},
  eprint        = {2602.03669},
  archivePrefix = {arXiv},
  primaryClass  = {cs.CV},
  doi           = {10.48550/arXiv.2602.03669}
}

@article{zou2019tvt,
  title={Robust lane detection from continuous driving scenes using deep neural networks},
  author={Q. Zou and H. Jiang and Q. Dai and Y. Yue and L. Chen and Q. Wang},
  journal={IEEE Transactions on Vehicular Technology},
  volume={69},
  number={1},
  pages={41--54},
  year={2020},
}

@article{abualsaud2021laneaf,
  author = {Abualsaud, Hala and Liu, Sean and Lu, David B. and Situ, Kenny and Rangesh, Akshay and Trivedi, Mohan M.},
  title = {LaneAF: Robust Multi-Lane Detection with Affinity Fields},
  journal = {IEEE Robotics and Automation Letters},
  volume = {6},
  number = {4},
  pages = {7477--7484},
  year = {2021}
}

@inproceedings{feng2022bezier,
  author = {Feng, Zhengyang and Guo, Shaohua and Tan, Xin and Xu, Ke and Wang, Min and Ma, Lizhuang},
  title = {Rethinking Efficient Lane Detection via Curve Modeling},
  booktitle = {Proceedings of the IEEE/CVF Conference on Computer Vision and Pattern Recognition},
  pages = {17062--17070},
  year = {2022}
}

@inproceedings{han2022laneformer,
  author = {Han, Jianhua and Deng, Xiajun and Cai, Xinyue and Yang, Zhen and Xu, Hang and Xu, Chunjing and Liang, Xiaodan},
  title = {Laneformer: Object-Aware Row-Column Transformers for Lane Detection},
  booktitle = {Proceedings of the AAAI Conference on Artificial Intelligence},
  volume = {36},
  pages = {799--807},
  year = {2022}
}

@inproceedings{he2016resnet,
  author = {He, Kaiming and Zhang, Xiangyu and Ren, Shaoqing and Sun, Jian},
  title = {Deep Residual Learning for Image Recognition},
  booktitle = {Proceedings of the IEEE Conference on Computer Vision and Pattern Recognition},
  pages = {770--778},
  year = {2016}
}

@inproceedings{honda2024clrernet,
  author = {Honda, Hiroto and Uchida, Yusuke},
  title = {CLRerNet: Improving Confidence of Lane Detection with LaneIoU},
  booktitle = {Proceedings of the IEEE/CVF Winter Conference on Applications of Computer Vision},
  pages = {1176--1185},
  year = {2024}
}

@inproceedings{hou2019sad,
  author = {Hou, Yuenan and Ma, Zheng and Liu, Chunxiao and Loy, Chen Change},
  title = {Learning Lightweight Lane Detection CNNs by Self Attention Distillation},
  booktitle = {Proceedings of the IEEE/CVF International Conference on Computer Vision},
  pages = {1013--1021},
  year = {2019}
}

@inproceedings{hu2018senet,
  author = {Hu, Jie and Shen, Li and Sun, Gang},
  title = {Squeeze-and-Excitation Networks},
  booktitle = {Proceedings of the IEEE Conference on Computer Vision and Pattern Recognition},
  pages = {7132--7141},
  year = {2018}
}

@inproceedings{jiang2018iounet,
  author = {Jiang, Borui and Luo, Ruixuan and Mao, Jiayuan and Xiao, Tete and Jiang, Yuning},
  title = {Acquisition of Localization Confidence for Accurate Object Detection},
  booktitle = {Proceedings of the European Conference on Computer Vision},
  pages = {784--799},
  year = {2018}
}

@inproceedings{li2020gfl,
  author = {Li, Xiang and Wang, Wenhai and Wu, Lijun and Chen, Shuo and Hu, Xiaolin and Li, Jun and Tang, Jinhui and Yang, Jian},
  title = {Generalized Focal Loss: Learning Qualified and Distributed Bounding Boxes for Dense Object Detection},
  booktitle = {Advances in Neural Information Processing Systems},
  volume = {33},
  pages = {21002--21012},
  year = {2020}
}

@inproceedings{liu2021condlanenet,
  author = {Liu, Lizhe and Chen, Xiaohao and Zhu, Siyu and Tan, Ping},
  title = {CondLaneNet: A Top-to-Down Lane Detection Framework Based on Conditional Convolution},
  booktitle = {Proceedings of the IEEE/CVF International Conference on Computer Vision},
  pages = {3773--3782},
  year = {2021}
}

@inproceedings{liu2021lstr,
  author = {Liu, Ruijin and Yuan, Zejian and Liu, Tie and Xiong, Zhiliang},
  title = {End-to-End Lane Shape Prediction with Transformers},
  booktitle = {Proceedings of the IEEE/CVF Winter Conference on Applications of Computer Vision},
  pages = {3694--3702},
  year = {2021}
}

@inproceedings{neven2018lanenet,
  author = {Neven, Davy and De Brabandere, Bert and Georgoulis, Stamatios and Proesmans, Marc and Van Gool, Luc},
  title = {Towards End-to-End Lane Detection: An Instance Segmentation Approach},
  booktitle = {IEEE Intelligent Vehicles Symposium},
  pages = {286--291},
  year = {2018}
}

@inproceedings{pan2018scnn,
  author = {Pan, Xingang and Shi, Jianping and Luo, Ping and Wang, Xiaogang and Tang, Xiaoou},
  title = {Spatial as Deep: Spatial CNN for Traffic Scene Understanding},
  booktitle = {Proceedings of the AAAI Conference on Artificial Intelligence},
  volume = {32},
  year = {2018}
}

@inproceedings{qin2020ufld,
  author = {Qin, Zequn and Wang, Huanyu and Li, Xi},
  title = {Ultra Fast Structure-Aware Deep Lane Detection},
  booktitle = {Proceedings of the European Conference on Computer Vision},
  pages = {276--291},
  year = {2020}
}

@article{qin2022ufldv2,
  author = {Qin, Zequn and Zhang, Pengyi and Li, Xi},
  title = {Ultra Fast Deep Lane Detection with Hybrid Anchor Driven Ordinal Classification},
  journal = {IEEE Transactions on Pattern Analysis and Machine Intelligence},
  pages = {1--14},
  year = {2022},
  doi = {10.1109/TPAMI.2022.3182097}
}

@inproceedings{qu2021fololane,
  author = {Qu, Zhan and Jin, Huan and Zhou, Yang and Yang, Zhen and Zhang, Wei},
  title = {Focus on Local: Detecting Lane Marker from Bottom Up via Key Point},
  booktitle = {Proceedings of the IEEE/CVF Conference on Computer Vision and Pattern Recognition},
  pages = {14122--14130},
  year = {2021}
}

@inproceedings{tabelini2020polylanenet,
  author = {Tabelini, Lucas and Berriel, Rodrigo and Paixao, Thiago M. and Badue, Claudine and De Souza, Alberto F. and Oliveira-Santos, Thiago},
  title = {PolyLaneNet: Lane Estimation via Deep Polynomial Regression},
  booktitle = {International Conference on Pattern Recognition},
  pages = {6150--6156},
  year = {2020}
}

@inproceedings{tabelini2021laneatt,
  author = {Tabelini, Lucas and Berriel, Rodrigo and Paixao, Thiago M. and Badue, Claudine and De Souza, Alberto F. and Oliveira-Santos, Thiago},
  title = {Keep Your Eyes on the Lane: Real-Time Attention-Guided Lane Detection},
  booktitle = {Proceedings of the IEEE/CVF Conference on Computer Vision and Pattern Recognition},
  pages = {294--302},
  year = {2021}
}

@misc{tusimple,
  author = {{TuSimple}},
  title = {TuSimple Lane Detection Benchmark},
  howpublished = {\url{https://github.com/TuSimple/tusimple-benchmark}},
  year = {2017}
}

@inproceedings{vaswani2017attention,
  author = {Vaswani, Ashish and Shazeer, Noam and Parmar, Niki and Uszkoreit, Jakob and Jones, Llion and Gomez, Aidan N. and Kaiser, Lukasz and Polosukhin, Illia},
  title = {Attention Is All You Need},
  booktitle = {Advances in Neural Information Processing Systems},
  volume = {30},
  year = {2017}
}

@inproceedings{wang2022ganet,
  author = {Wang, Jinsheng and Ma, Yinchao and Huang, Shaofei and Hui, Tianrui and Wang, Fei and Qian, Chen and Zhang, Tianzhu},
  title = {A Keypoint-Based Global Association Network for Lane Detection},
  booktitle = {Proceedings of the IEEE/CVF Conference on Computer Vision and Pattern Recognition},
  pages = {1392--1401},
  year = {2022}
}

@inproceedings{woo2018cbam,
  author = {Woo, Sanghyun and Park, Jongchan and Lee, Joon-Young and Kweon, In So},
  title = {CBAM: Convolutional Block Attention Module},
  booktitle = {Proceedings of the European Conference on Computer Vision},
  pages = {3--19},
  year = {2018}
}

@inproceedings{xiao2023adnet,
  author = {Xiao, Lingyu and Li, Xiang and Yang, Sen and Yang, Wankou},
  title = {ADNet: Lane Shape Prediction via Anchor Decomposition},
  booktitle = {Proceedings of the IEEE/CVF International Conference on Computer Vision},
  pages = {6404--6413},
  year = {2023}
}

@inproceedings{zhang2021varifocalnet,
  author = {Zhang, Haoyang and Wang, Ying and Dayoub, Feras and Sunderhauf, Niko},
  title = {VarifocalNet: An IoU-Aware Dense Object Detector},
  booktitle = {Proceedings of the IEEE/CVF Conference on Computer Vision and Pattern Recognition},
  pages = {8514--8523},
  year = {2021}
}

@inproceedings{zhang2021vil100,
  author = {Zhang, Yujun and Zhu, Lei and Feng, Wei and Fu, Huazhu and Wang, Mingqian and Li, Qingxia and Li, Cheng and Wang, Song},
  title = {VIL-100: A New Dataset and A Baseline Model for Video Instance Lane Detection},
  booktitle = {Proceedings of the IEEE/CVF International Conference on Computer Vision},
  pages = {15681--15690},
  year = {2021}
}

@inproceedings{zhang2022mhvt,
  author = {Zhang, Han and Gu, Yunchao and Wang, Xinliang and Pan, Junjun and Wang, Minghui},
  title = {Lane Detection Transformer Based on Multi-frame Horizontal and Vertical Attention and Visual Transformer Module},
  booktitle = {Proceedings of the European Conference on Computer Vision},
  pages = {1--16},
  year = {2022},
  doi = {10.1007/978-3-031-19842-7_1}
}

@inproceedings{zheng2021resa,
  author = {Zheng, Tu and Fang, Hao and Zhang, Yi and Tang, Wenjian and Yang, Zheng and Liu, Haifeng and Cai, Deng},
  title = {RESA: Recurrent Feature-Shift Aggregator for Lane Detection},
  booktitle = {Proceedings of the AAAI Conference on Artificial Intelligence},
  volume = {35},
  pages = {3547--3554},
  year = {2021}
}

@inproceedings{zheng2022clrnet,
  author = {Zheng, Tu and Huang, Yifei and Liu, Yang and Tang, Wenjian and Yang, Zheng and Cai, Deng and He, Xiaofei},
  title = {CLRNet: Cross Layer Refinement Network for Lane Detection},
  booktitle = {Proceedings of the IEEE/CVF Conference on Computer Vision and Pattern Recognition},
  pages = {898--907},
  year = {2022}
}
\end{document}